\documentclass[conference]{IEEEtran}
\usepackage{enumerate}
\usepackage{booktabs}
\usepackage{amssymb}
\usepackage{color,hyperref}
\usepackage{multirow}
\usepackage{algorithm}
\usepackage{algpseudocode}
\usepackage{comment}
\usepackage{amsmath}
\usepackage{bm}
\usepackage{latexsym}
\usepackage{graphicx}
\usepackage{subfigure}
\usepackage{epstopdf}
\usepackage{epsfig}
\usepackage{pifont}
\usepackage{xspace}
\usepackage[bottom]{footmisc}
\usepackage[11pt]{moresize}
\usepackage{todonotes}
\usepackage{url}
\newcommand{\mypara}[1]{
	\vspace*{0.01cm}
	\noindent\textbf{\textit{#1}}}
\newcommand{\RNum}[1]{\uppercase\expandafter{\romannumeral #1\relax}}
\DeclareMathAlphabet{\mathpzc}{OT1}{pzc}{m}{it}
\newcommand{\model}{\textbf{\texttt{HDXplore}}}

\begin{document}
\title{HDXplore: Automated Blackbox Testing of Brain-Inspired Hyperdimensional Computing}
\author{
	Rahul Thapa, Dongning Ma, Xun Jiao \\
	Villanova University \\
	\{rthapa, dma2, xjiao\}@villanova.edu
}
\maketitle

\begin{abstract}
Inspired by the way human brain works, the emerging hyperdimensional computing (HDC) is getting more and more attention. 
HDC is an emerging computing scheme based on the working mechanism of brain that computes with deep and abstract patterns of neural activity instead of actual numbers. Compared with traditional ML algorithms such as DNN, HDC is more memory-centric, granting it advantages such as relatively smaller model size, less computation cost, and one-shot learning, making it a promising candidate in low-cost computing platforms. However, the robustness of HDC models have not been systematically studied. In this paper, we systematically expose the unexpected or incorrect behaviors of HDC models by developing \model, a blackbox differential testing-based framework. We leverage multiple HDC models with similar functionality as cross-referencing oracles to avoid manual checking or labeling the original input. We also propose different perturbation mechanisms in \model. \model ~automatically finds thousands of incorrect corner case behaviors of the HDC model. We propose two retraining mechanisms and using the corner cases generated by \model ~to retrain the HDC model, we can improve the model accuracy by up to 9\%. 
\end{abstract}

\section{Introduction}
Recently, inspired by the way human brain works, hyperdimensional computing (HDC), also known as vector-symbolic architectures, is emerging as a viable alternative to DNNs. Instead of actual numbers, HDC computes with deep and abstract patterns of neural activity similar to human brain. Compared with deep neural networks (DNNs), HDC is more memory-centric, granting it advantages such as relatively smaller model size, less computation cost, and one-shot learning capability~\cite{ge2020classification}. Recently, HDC has demonstrated promising results in various problem domains including language classification~\cite{rahimi2016robust}, brain computer interfaces~\cite{rahimi2017hyperdimensional}, DNA pattern matching~\cite{kim2020geniehd}, and anomaly detection~\cite{wang2021hdad}. However, despite the growing popularity of HDC, the robustness of HDC models has yet been systematically studied. 

Testing is an inherently important step in enhancing the robustness of systems or programs. 
Conventionally, testing a ML system often requires a large-scaled of manually labeled datasets and feed them to the ML systems for testing. Then, the mis-classified samples will be used to retrain or fine-tune the ML systems to improve the accuracy. Such approach, however, is hardly feasible and scalable any longer as the ML systems are scaling significantly to a more sophisticated input space. The challenges for ML system testing is further exacerbated by the fact that ML systems are found to be vulnerable to invisible perturbations to original inputs. That is, ML systems can be ``fooled'' and produce wrong predictions~\cite{szegedy2013intriguing}. 
In observing such phenomenon, this paper aims to answer the following important question: How can we automatically and systematically test HDC systems to detect and fix potential flaws or undesired behaviors.

The key challenges in automated systematic testing of HDC systems are twofold: (i) Large-scale manual labeling: As mentioned, conventional ML testing requires the gathering of a large set of data samples to be feed into ML systems that is usually unscalable and infeasible~\cite{russakovsky2015imagenet, wang2019exploring}. Google even used simulation to generate synthetic data~\cite{madrigal2017inside}. Note that such manual effort is not only short of scalability but also is largely randomized, making it unable to cover more than a tiny fraction of all possible corner cases. (ii) Lack of mathematical architecture: Unlike ML methods such as DNNs with a well-defined mathematical formulation, HDC largely relies on random project-based encoding (as explained later)~\cite{rahimi2016robust}, adding difficulty to efficiently acquire adequate information to guide the testing process. As a result, difference-inducing images generation techniques used in DNNs cannot be applied here since they rely on a set of well-defined mathematical optimization problems~\cite{goodfellow2014explaining}. 

\begin{figure}
	\centering
	\includegraphics[page=1, width=1\columnwidth]{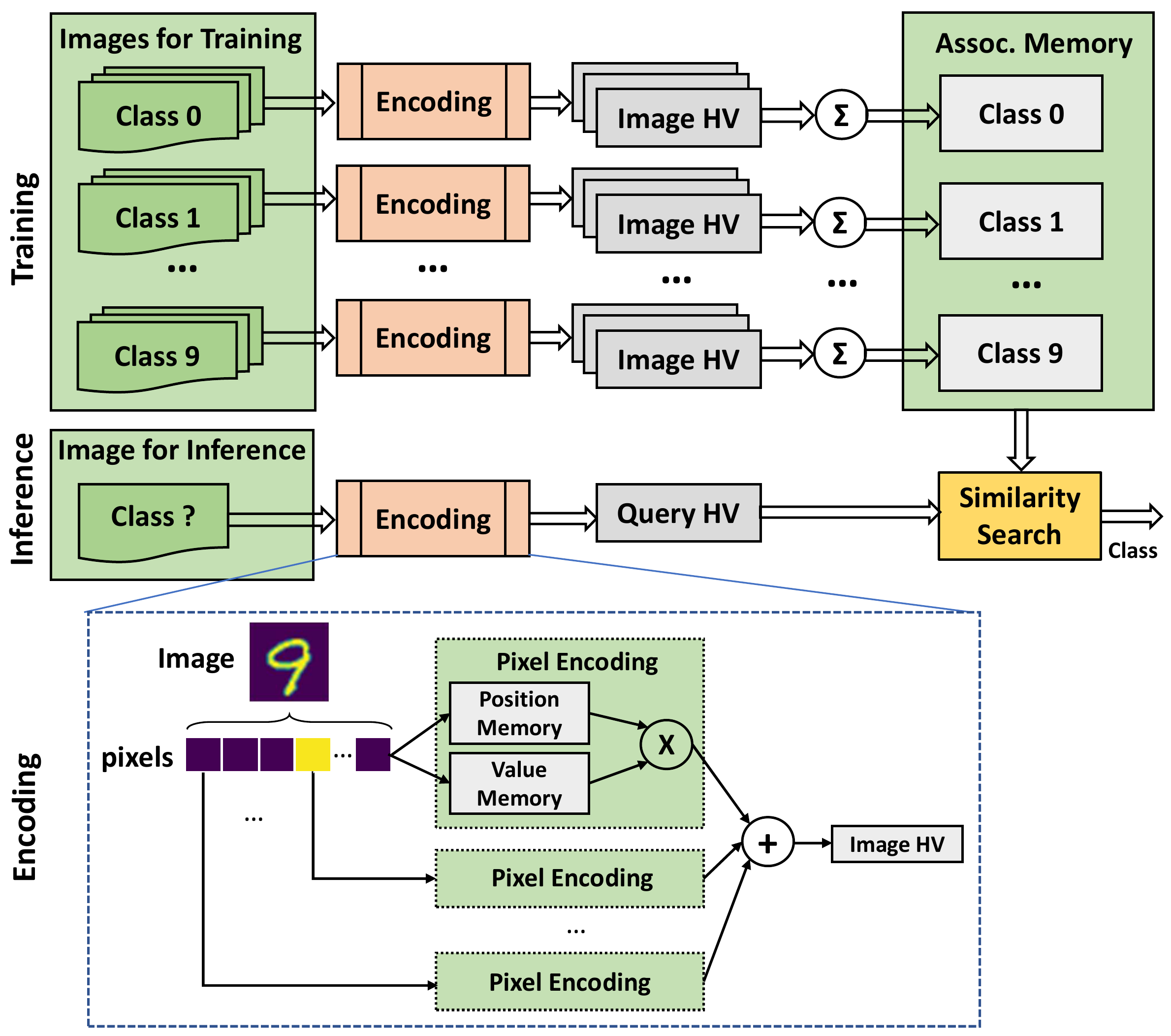}
	\caption{HDC for image classification with three phases: (1) \textbf{Encoding}, (2) \textbf{Training} and (3) \textbf{Inference}.}
	\label{fig:hdc}
	\vspace{-0.6cm}
\end{figure}

To address such challenges, we introduce \model, a highly-automated testing framework for HDC based on differential testing. Differential testing, also known as differential fuzzing, is a popular software testing technique that attempts to detect bugs, by providing the same input to different implementations of the same application, and observing differences in their execution~\cite{chen2016coverage}. We assume a blackbox testing scenario as we assume we have no knowledge of the internal HDC architecture. 
We make the following contributions: 
\begin{itemize}
	\item 
	We develop \model, the first blackbox testing framework that can automatically expose the incorrect behaviors of HDC classifiers in a highly-automated and scalable way without manual labeling. 
	\item 
	We develop two modes for \model~ --- the original mode and the perturbation mode. In the original mode, we construct multiple HDC classifiers and then use \model ~to cross-reference each other to identify differential behaviors. In the perturbation mode, we perturb images and use differential testing to automatically find incorrect behaviors. 
	\item 
	Experimental results on MNIST dataset show that \model ~can efficiently find a large number of incorrect corner cases in HDC classifiers. Using the \model-generated corner cases, we develop two retraining methods, static retraining and dynamic retraining. 
	The retraining methods can improve the HDC classifier accuracy by up to 9\%. 
\end{itemize}

\section{Related Work}
Rahimi et al. used HDC on hand gesture recognition and achieves 97.8\% accuracy on average , which surpassed the support vector machine by 8.1\%~\cite{rahimi2016hyperdimensional}. HDC was also used for language classification with an accuracy at 97\%~\cite{rahimi2016robust}. %They further combines HDC with NN to establish \textit{VoiceHD-NN} for accuracy improvement. 
Manabat et al., applied HDC to character recognition and conducted performance analysis~\cite{manabat2019performance}. For optimization of HDC processing, \textit{HDC-IM}~\cite{liu2019hdc} proposed in-memory computing techniques for HDC scenarios based on Resistive Random-Access Memory (RRAM). There are also optimizations on HDC targeted at different computing platforms such as the FPGA~\cite{schmuck2019hardware} and the 3D IC~\cite{wu2018brain}.  

Recently, the emerging adversarial attacks on deep learning systems~\cite{goodfellow2014explaining, nguyen2015deep}
have demonstrated that even the most advanced DNNs can be fooled by applying invisible perturbations. Most of adversarial images are generated by leveraging the mathematical properties of DNNs. For example, a fast gradient sign method was developed with required gradient computed efficiently using back-propagation~\cite{goodfellow2014explaining}. The gradient of the posterior probability for a specific class (e.g., softmax output) was computed with respect to the input image using back-propagation, which was then used to increase a chosen unit’s activation to obtain adversarial images~\cite{nguyen2015deep}. 
However, these methods cannot be applied to HDC because HDC is not built upon solving a mathematical optimization problem to find hyperparameters. Further, usually the testing of DNNs and HDC models assumes a greybox or even whitebox scenario where the testers have the knowledge of internal model strucure~\cite{ma2021hdtest, goodfellow2014explaining, nguyen2015deep}. The closest study to our work is HDTest~\cite{ma2021hdtest} but it assumes a whitebox testing scenario. In contrast, this paper presents the first blackbox testing framework for HDC models.

\section{HDC for Image Classification}
\label{sec:hdc}
We develop an HDC classifier for image classification as our testing target. There are three key phases in HDC models: \textbf{Encoding}, \textbf{Training} and \textbf{Inference} as illustrated in Fig.~\ref{fig:hdc}.

\subsection{Encoding}
The encoding phase is to use HDC arithmetic to encode an image into a hypervector (HV) called ``Image HV''. HV is the fundamental building block of HDC. They are high-dimensional, holographic, and (pseudo-)random with independent and identically distributed (i.i.d.) components. In HDC, HV supports three types of arithmetic operation for encoding: (element-wise) addition, (element-wise) multiplication and permutation (cyclic shifting). Multiplication and permutation will produce HVs that are orthogonal to the original operand HVs while addition will preserve 50\% of each original operand HVs~\cite{rahimi2016hyperdimensional}.

As shown in Fig.~\ref{fig:hdc}(1), in order to encode one image into its representing HV, there are three steps. The first step is to decompose and flat the image into an array of pixels. The indices of the pixels in the array reflect the position of the pixel in the original image while the values of the pixels reflect the greyscale level of each pixel. For the MNIST dataset we use in this paper, since the image size is $28 \times 28$ and the pixel range is 0 to 255 in greyscale, we flat a single image into an array with 784 elements with values ranging from 0 to 255. 

The second step is to construct HVs representing each pixel from the index and value information provided by the image array. We randomly generate two memories of HVs: the position HV memory and the value HV memory based on the size and pixel value range of the image. The position HV memory accommodates $28 \times 28 = 784$ HVs, each representing a pixel's position in the original image. The value HV memory accommodates $255$ HVs, each representing a pixel's greyscale value. For each pixel from the image array, we look up the position HV and value HV from the two memories and use multiplication operation to combine them in order to encode the pixel into its representing HV. For example, if the pixel's index is 128 and the greyscale value is 192, its representing HV is obtained by: $PixelHV = PositionMem[128] \circledast ValueMem[192]$.

The third step is to establish the HV representing the entire image. After encoding all the 784 pixels into pixel HVs, the final image HV is established by summing up the pixel HVs. So far, we encode one image into a representing image HV that is ready for HDC training and testing as one sample. %However, addition can destroy the bipolar distribution of the HV (some elements can be numbers other than ``1'' or ``-1''), thus the image HV is bipolarized again by Eq.~\ref{eqn:bipolar}. 

\subsection{Training}
The training phase is to iteratively incorporate the information contained in each image in the training set into the associative memory (AM) with corresponding label. AM stores a group of HVs, each representing a class. In other words, training is the process of building the AM by adding up all the training images' HVs belong to one class together. In the beginning, we initialize every class HV inside the AM with zero. Note that the dimension of the class HVs inside the AM is consistent with the image HVs. Therefore we can add every image HV into the corresponding class HV according to the label, to train the AM. %Then, for each image with class label $c_i$ in the training set $T$, we first encode the image into its representing HV $HV_i$ and then we add $HV_i$ into the corresponding class HV inside the associative memory $AM[c_i]$. After one epoch when all the training images are added into $AM$, we finally bipolarize the HVs inside $AM$ again using Eq.\ref{eqn:bipolar} and the $AM$ is constructed and ready for testing and evaluation.

\subsection{Inference}
The inference phase is to evaluate the trained AM using the unseen inference dataset. First, every image for inference is encoded into its representing (query) image HV, using the identical encoding mechanism and generated position and value memory in the training phase. Then we calculate the similarity between the query HV and every class HV inside the AM. In \model~, similarity is measured using cosine similarity in Eq.~\ref{eqn:cosim}: 
\begin{equation}
    Sim = CoSim(qHV, AM[i]) = \frac{qHV \cdot AM[i]}{||qHV|| ||AM[i]||}
    \label{eqn:cosim}
\end{equation}

where $qHV$ refers to the query image HV and $AM[i]$ refers to the i-th class HV inside AM. The class with the maximum similarity with the query image HV subsequently becomes the prediction result of this image. We then compare the true label with the prediction for this image to determine if the prediction is correct. We iterate all the images in the inference test to evaluate the accuracy of the HDC classifier.

\section{\model ~Framework}
\begin{figure}[htbp]
    \centering
    \includegraphics[keepaspectratio, width = 1\columnwidth]{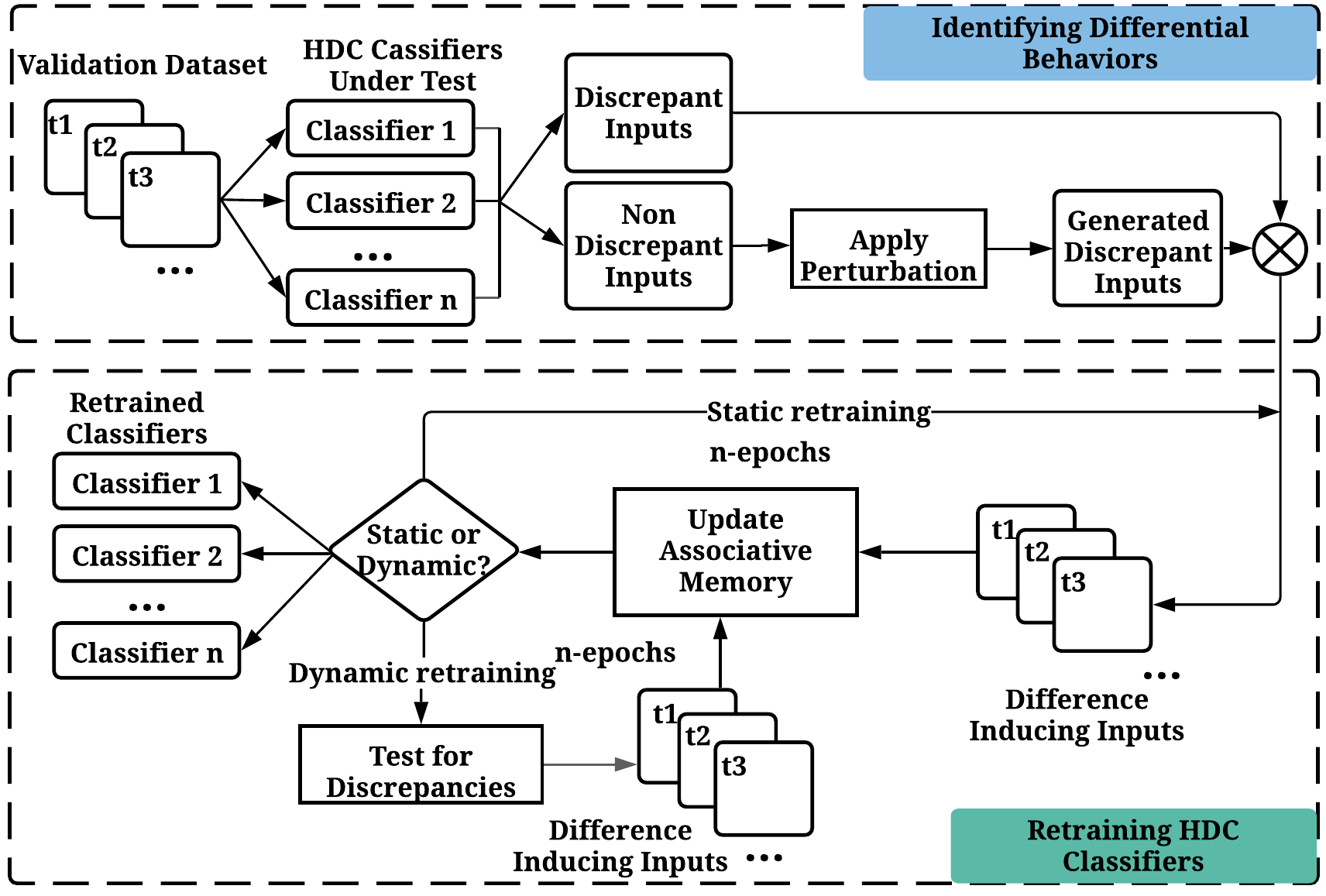}
    \caption{Overview of \model ~with two phases: phase 1 use differential testing to identify difference-inducing inputs, and phase 2 use difference-inducing inputs to retrain the HDC classifiers.}
    \label{fig: HDXplore_workflow}
\end{figure}

\begin{figure}[htbp]
    \centering
    \subfigure[Same prediction]{  
        \includegraphics[keepaspectratio,width=0.35\columnwidth]{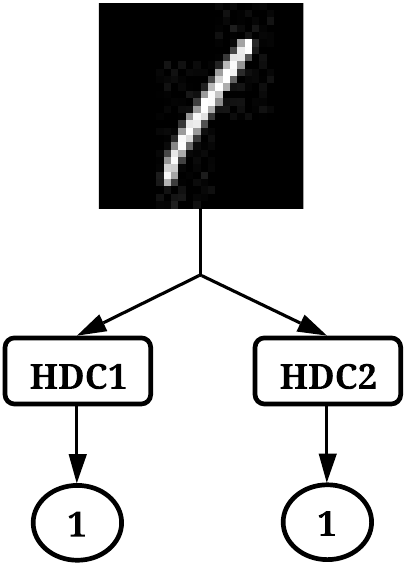}
    }
    \subfigure[Difference Inducing prediction]{
        \includegraphics[keepaspectratio,width=0.35\columnwidth]{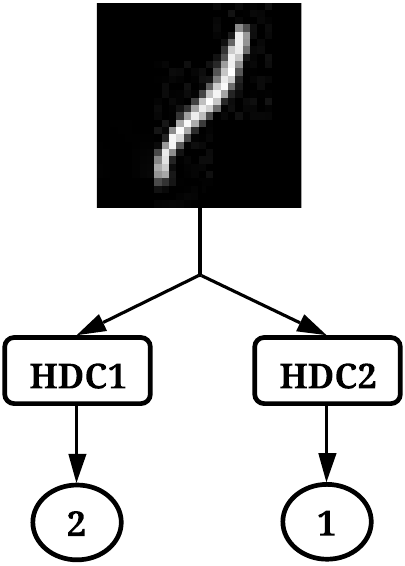}
    }
    \vspace{-0.2cm}
    \caption{The first image is predicted as "1" by both HDC classifiers. The second image is predicted differently by two classifiers, hence a difference inducing is found.}
    \label{fig:discrepancies_HDC}
    \vspace{-0.2cm}
\end{figure}

In this section, we describe \model ~framework for systematically testing HDCs to locate erroneous corner case behaviors on the image classification task. The main components of \model ~are shown in Fig.~\ref{fig: HDXplore_workflow}. \model ~takes $N$ trained classifiers as well as validation dataset. The MNIST dataset is divided into training, validation, and testing set. We split the 60,000 MNIST training dataset into training and validation set. We do this because for a testing problem, it is natural to assume that we are given already-trained classifiers. Hence, we use the training dataset to train the initial HDC classifiers first. Then, we use validation dataset for \model. \model ~uses validation set to generate new corner cases that cause the $N$ HDC classifiers to behave inconsistently, i.e., the HDC classifiers are producing different predictions. We refer to such cases as difference-inducing inputs. The identified difference-inducing inputs will then be used to retrain the classifier to improve accuracy and robustness. The remaining 10,000 completely unseen testing dataset is preserved for the ultimate evaluation of the performance of HDC classifier after going through \model. 

We use Fig.~\ref{fig:discrepancies_HDC} as an example to show how \model ~generates test inputs. Consider that we have two HDCs to test, both of which perform similar tasks, i.e. classifying digits from 1-9. They are trained on the same dataset but independently with different randomly generated HDC parameters, e.g., position and value HVs. Therefore, although the HDCs uses the same encoding, training and testing scheme, they have slightly different classification rules. Assume we have an image which both HDCs identify as a ``1'', as shown in Fig.~\ref{fig:discrepancies_HDC} (a), \model ~tries to find differential behaviors by perturbing the input which will lead to different HDC classification outputs, e.g., one HDC classifier classifies it as a ``1'' while the other classifiers as a ``2'', as shown in Fig.~\ref{fig:discrepancies_HDC} (b).

\begin{algorithm}
\small
    \caption{\model}
    \algrenewcommand\algorithmicrequire{\textbf{Input}}
    \algrenewcommand\algorithmicensure{\textbf{Output}}
    \begin{algorithmic}[1]
    \Require inputs $\gets$ validation dataset; classifiers $\gets$ multiple HDCs under test; seeds $\gets$ random seeds for HVs; perturbations $\gets$ predefined perturbation functions; epochs $\gets$ number of epochs for retraining
    \Ensure retrained classifiers, difference-inducing images $C$
    \State /* \emph{main procedure} */
    \State gen\_test $\gets$ empty\_set
    \State /* \emph{finds images that all classifiers disagree and agree on respectively} */
    \State dis\_images, non\_dis\_images $\gets$ discrepancies(classifiers, inputs)
    \For{image $\in$ non\_dis\_images}
    \State /* \emph{randomly picks one of the classifiers} */
    \State m $\gets$ random(classifiers)
    \For{perturb $\in$ perturbations}
    \State /* \emph{applies the perturbation function to the image} */
    \State n\_image = perturb(image)
    \If{m.predict(n\_image) $\neq$ (classifiers-m).predict(n\_image)}
    \State /* \emph{classifiers predict n\_image differently} */
    \State gen\_test.add(n\_image)
    \EndIf
    \EndFor
    \EndFor
    \State dis\_images.add(gen\_test)
    \State /* \emph{retrains the classifier for given epochs} */
    \For{epoch $\in$ epochs}
    \For{image $\in$ dis\_images}
    \For{seed $\in$ seeds}
    \State /* \emph{finds true and wrong label for given image} */
    \State y\_wrong $\gets$ get\_wrong\_label(image, classifiers[seed])
    \State y\_true $\gets$ get\_true\_label(image)
    \State /* \emph{gets HV for the given image} */
    \State hv $\gets$ projection(image, seed)
    \State /* \emph{subtracts the HV from wrong label and adds to correct label} */
    \State classifiers[seed][y\_wrong] -$\gets$ hv
    \State classifiers[seed][y\_true] +$\gets$ hv
    \EndFor
    \EndFor
    \State dis\_images, non\_dis\_images $\gets$ discrepancies(classifiers, dis\_images)
    \EndFor \\
    \Return classifiers, gen\_test
    \end{algorithmic}
    \label{alg:HDXplore}

\end{algorithm}

\mypara{Identifying Difference-Inducing Inputs (Line 3 - 17)}: Algorithm \ref{alg:HDXplore} shows the algorithm of \model ~for identifying difference-inducing images and using those inputs to retrain the HDC classifiers. The first objective of the algorithm is to identify inputs that can induce inconsistent behaviors in the HDCs, i.e., different HDC classifiers will classify the same input into different classes. Suppose we have $n$ HDC classifiers $H_{k \in 1..n}$ : x $\rightarrow$ y, where $H_k$ is the function modeled by the $k-$th HDC classifier. $x$ represents the input and $y$ represents the output class probability vectors. If an arbitrary $x$ gets classified to a different class by at least one HDC classifiers, then such input is flagged as difference-inducing input $x_d$, and subsequently separated from the set of original images. By the end of this process, what remains is a set of images of which the classification is consistent across all HDC classifiers. Next, the goal of \model ~is to perturb an image $x\_$ from this set into $x\_'$, trying to make its classification across all the $n$ HDC classifiers inconsistent. For perturbing $x$, we use four predefined perturbation functions (as detailed later). The resulting difference-inducing images $x\_'$ are combined with the initial difference-inducing images in the validation set $x_d$, to form a complete set of difference-inducing images.

\mypara{Retraining HDC Classifiers (Line 18 -31)}: After generating the difference-inducing images, the second objective of \model ~algorithm is to use these images to retrain the HDC classifiers to improve its robustness and generalizability. Note that in this retraining process, we would need to label the difference-inducing images, which is significantly fewer than labeling the original images. We develop two retraining methods: static retraining and dynamic retraining. 

For static retraining, we use the same set of difference-inducing images obtained in the first epoch to iteratively retrain the HDC classifier for multiple epoches. By retraining, we mean that we update the HVs in the associative memory for every subsequent epoch because the HVs are the final trained parameters of HDC classifiers. The retraining process is as follows: For each difference-inducing image, its image HV is subtracted from the wrong class HV in the associate memory and added to the correct class HV in the associative memory. The static retraining, however, always use the same set of difference-inducing images to update the HVs. This does not consider the fact that, during retraining, the difference-inducing images may change for HDC classifiers. 

Hence, we propose dynamic retraining, where we will update the difference-inducing images in every epoch. That is, we use the \model ~to generate a new set of difference-inducing images in every retraining epoch and use them to update the HVs in the associate memory. The update process is still by subtracting image HVs from the wrong class HV and adding image HV to the correct class HV.

\section{Experimental Results}
\subsection{Experimental Setup}
We use MNIST \cite{lecun1998mnist} dataset and evaluate \model ~on a group of three HDCs. The MNIST dataset is divided into 3 sets: the training, validation, and testing set. Originally, MNIST has 60,000 training images and 10,000 testing images. We split the training set, i.e., 60,000 images, into a training set (e.g., 30,000 images) and a validation set (e.g., the remaining 30,000 images). As a testing problem, we assume classifiers are pre-trained. Thus, we train three classifiers using only the training set (e.g, 30,000 images) and use \model ~only on the validation set (e.g., the remaining 30,000 images). Finally, the \model~generated retrained classifier will be evaluated using the completely unseen 10,000 images in the test set. We have explored 5 different split ratios between training and validation datasets. However, we only present the results for the best performing split, i.e. 50\%. By 50\% (X\%) split, we mean 50\% (X\%) of entire training datasets are used as validation dataset. 

We apply four perturbations to the images to create difference-inducing images: skew, noise, brightness, and elastic transform. To skew an image, we randomly initialize a distribution with an off-center mean %(12) 
and non-zero standard deviation %(1) 
and distort all pixels by that amount. Similarly, to add noise to the image, we randomly choose 100 points in the images and overwrite the pixel value to 0. We change the brightness by simply decreasing the pixel values by a given factor. % (0.8). 
Finally, for elastic transform, we replicate the elastic deformation of images as described in \cite{simard2003best}.

\begin{figure}[htbp]

    \centering
    \subfigure[Original Image]{
        \includegraphics[keepaspectratio,width=0.9\columnwidth]{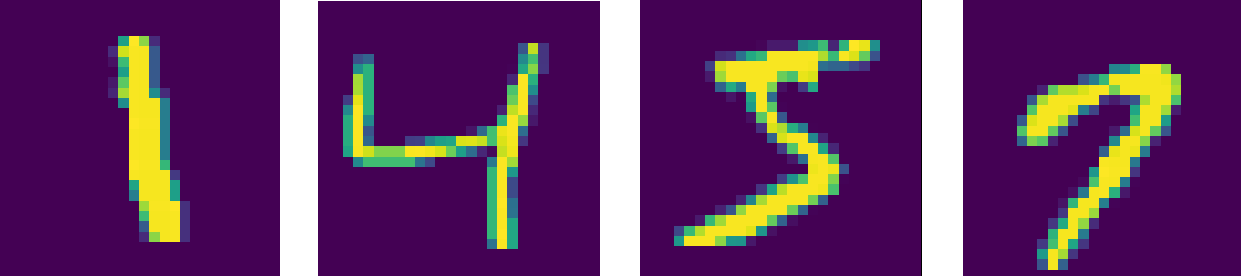}
    }
    \subfigure[Generated (adversarial) images]{  
        \includegraphics[keepaspectratio,width=0.9\columnwidth]{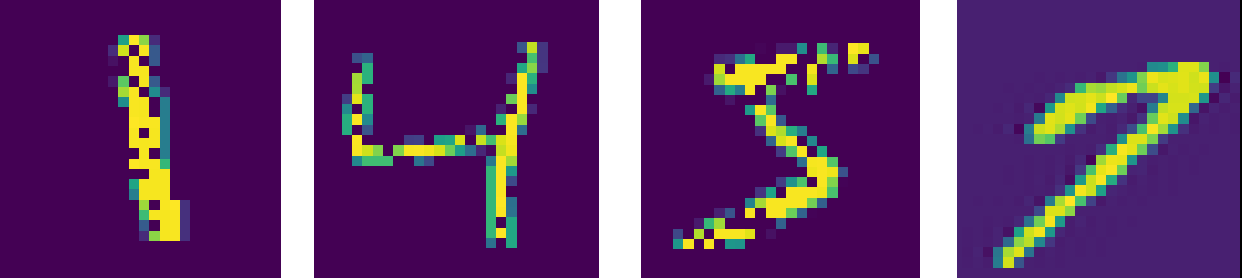}
    }
    \caption{Sample adversarial images produced by \model.}
    \label{fig:original_perturbed_images}

\end{figure}

\begin{figure*}[htbp]

    \centering
    \subfigure[Dynamic Retraining]{
        \includegraphics[keepaspectratio,width=0.9\columnwidth]{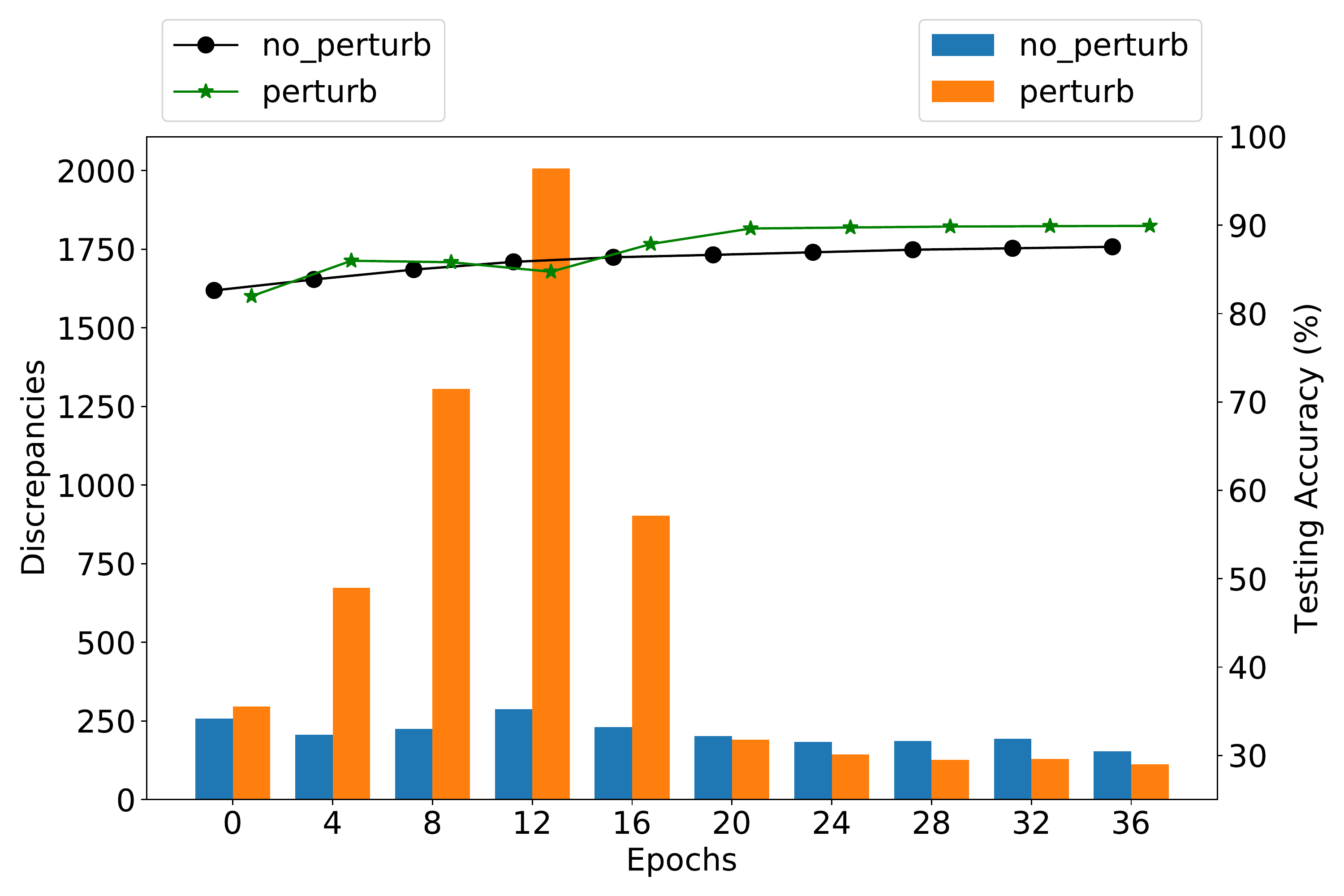}
    }
    \subfigure[Static Retraining]{  
        \includegraphics[keepaspectratio,width=0.9\columnwidth]{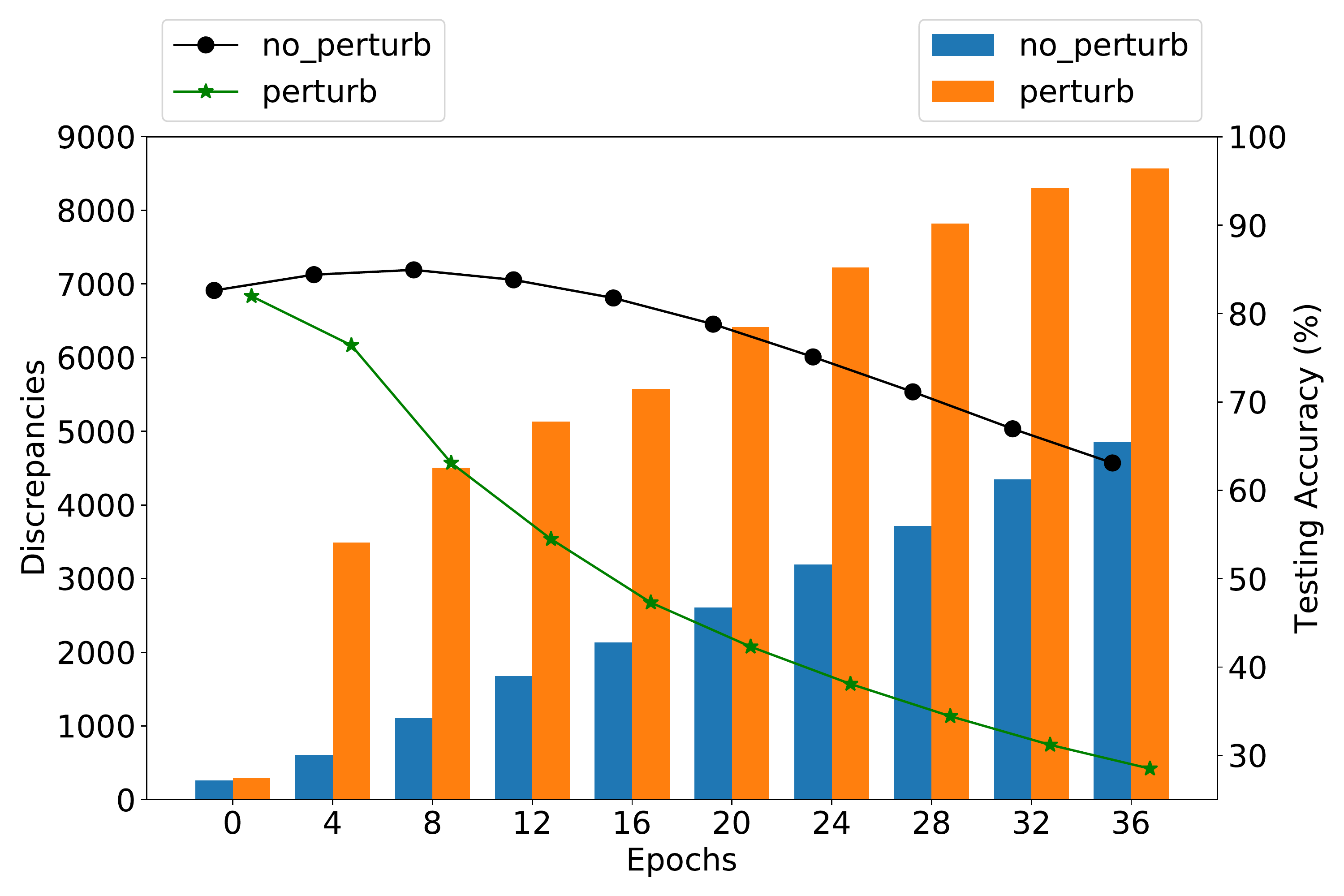}
    }
    \caption{Accuracy and number of discrepancies of \model ~retrained classifiers on unseen 10000 MNIST images for each epoch of retraining process. We conducted 4 different experiments with the combination of 2 different parameters. (1) Static vs. Dynamic; (2) Perturbation vs. No-perturbation.}
    \label{fig:accuracy_discrepancies_curve}

\end{figure*}

\subsection{Difference-Inducing Images}
For initially training HDC classifiers, 3 different random seeds (30, 40, and 50) are used to generate random base HVs.  
The individual images are then encoded using the base HVs and are added to the corresponding HVs in the associative memory. All 3 HDC classifiers achieve $\sim$81\% accuracy. \model ~is able to find 966 difference-inducing images in the validation set initially. After perturbation, \model ~generated additional around 4000 difference-inducing images. On average, \model ~generates 3-4 difference-inducing images per sec on the MNIST dataset for all splits. We run \model ~algorithm for 40 epochs for all our experiments and on each epoch, we evaluate the classification accuracy and number of difference-inducing images using 10,000 unseen testing images.
Fig.~\ref{fig:original_perturbed_images} presents several generated examples of adversarial images.

\subsection{Static Vs. Dynamic Retraining}
Through our dynamic retraining method, we see an overall increase in accuracy from ~81\% up to 90\% over the course of 40 epochs as shown in Fig.~\ref{fig:accuracy_discrepancies_curve} (a). For static retraining in Fig.~\ref{fig:accuracy_discrepancies_curve} (b), even though the accuracy seems to increase in the beginning few epochs, the accuracy eventually starts dropping continuously and significantly for all settings.  

This is consistent with the number of difference-inducing images as shown in Fig.~\ref{fig:accuracy_discrepancies_curve}. With dynamic retraining, there is an overall decrease in the number of difference-inducing images, decreasing from 296 to 153 for ``no perturbation'' and from 296 to 104 for ``perturbation''. With static retraining, however, we see that the number of difference-inducing images is strictly increasing in all settings.

Comparing the accuracy and discrepancy curves, we see that for dynamic retraining, the decrease in the number of difference-inducing images is simultaneously followed by an increase in the accuracy. For static retraining, however, the trend is not as clear, as we can see that the accuracy increases in the beginning few epochs even when the difference-inducing images were increasing. Overall, the dynamic retraining method seems to be increasing the accuracy of HDC classifiers while simultaneously decreasing the number of difference-inducing images, making the classifiers both accurate and robust. 

There is a drop in the accuracy (from $\sim$86\% to $\sim$82\%) and spike in the number of difference-inducing images (from $\sim$500 to $\sim$2000) for dynamic retraining with perturbation from 5 to 11 epochs. 

For dynamic retraining, this means that the HDC classifiers get to retrain on a higher number of difference-inducing images while updating the difference-inducing set on each epoch. However, for static retraining, the same large number of initial difference-inducing images are used to retrain the classifiers over and over again. Therefore, the difference-inducing images get compounded on each epoch without getting a chance to improve upon them, eventually resulting in a drop in accuracy. 
Even for dynamic retraining, we noticed a drastic drop in accuracy while applying perturbation from 4 to 11 epochs. 
This happened because the number of difference-inducing images kept on increasing from the very beginning. Therefore, even if the classifier's performance improved in the beginning epochs, it dropped drastically as the number of difference-inducing images significantly increased. However, because dynamic retraining updates the number of difference-inducing images on each epoch, the classifiers eventually start learning from all these difference-inducing images resulting in a drop in the number of difference-inducing images and a significant increase in the accuracy.

\subsection{Perturbation Vs. Non-Perturbation}
Applying perturbation results in an overall greater increase in accuracy in less number of epochs than without perturbation in dynamic retraining. We see that the highest accuracy improvement (from ~81\% to ~90\%) is seen when the perturbation is applied. However, for static retraining, the case is slightly different, where for the same split, applying perturbation results in a significant and rapid decrease in overall accuracy than without perturbation. The reason can be seen in discrepancy bar in Fig.~\ref{fig:accuracy_discrepancies_curve} (b). Applying perturbation increases the number of difference-inducing images in the validation set significantly. Even though the accuracy of the classifiers seems to improve in the beginning epochs, as the number of difference-inducing images starts increasing rapidly, the accuracy of the classifiers starts to drop simultaneously with the number of difference-inducing images in the validation dataset. Comparing the accuracy and discrepancy plot in Fig.~\ref{fig:accuracy_discrepancies_curve} (b), we notice that higher the difference-inducing images in the validation dataset, the sooner and faster the accuracy drops for static retraining.

Retraining the classifiers dynamically using all the difference-inducing images found after applying perturbation also increases the robustness of the classifiers. To quantify the robustness, we used same as well as slightly different perturbations that we used in retraining our HDC classifiers in all the images from testing dataset that all classifiers initially agreed on. We then compared the number of difference-inducing images that our framework was able to produce. We found $\sim$1200 difference-inducing images in the the baseline classifiers, i.e. the classifier without any retraining. Similarly, we found $\sim$800 difference-inducing images in the classifiers that was retrained but without any perturbed images. However, we found $\sim$600 difference inducing images for our dynamically retrained classifiers with perturbed images found through our framework. Based on this metric, our framework made HDC classifiers about 50\% more robust than our baseline model and 25\% more robust than the dynamically retrained model without the perturbed images found through our framework.

\subsection{Exploration on Varying Split Ratios}
We additionally experimented with various split ratios in addition to 50\%. We found a general trend that seems to apply for most of the splits. For split ratio smaller than 50\%, we noticed that the accuracy generally increases and number of discrepancies generally decreases for dynamic retraining. For static retraining, the case is reverse, consistent with the results from 50\% split. However, the case is a little bit different when you split the data too high above 50\% (say 60\%). In this case, even for dynamic retraining, with or without perturbation, the accuracy decreases and the discrepancy increases significantly. Unlike with 50\% split, where the classifier eventually starts learning from all the discrepancies after a certain number of epochs, with splits significantly higher that 50\%, the classifier never recovers.

\section{Conclusion}
This paper presents \model, a highly-automated and scalable blackbox testing approach for HDC models. Based on differential testing, \model ~iteratively mutates inputs to generate difference-inducing images to expose incorrect behaviors of HDC models. We develop multiple modes in \model, and evaluate \model ~on MNIST dataset. Experimental results show that \model ~is able to generate difference-inducing images efficiently with and without perturbations. We use the \model-generated inputs to retrain HDC classifiers under static and dynamic retraining, which can further improve the accuracy and robustness of HDC classifiers. While HDC performance still expects more advancements both in theoretical and implementation aspects, this paper aims to shed light on the robustness aspects of this emerging technique.

\bibliography{HDXplore}

\end{document}